\documentclass{article}

\usepackage[final]{neurips_2025}
\usepackage[utf8]{inputenc}
\usepackage[T1]{fontenc}
\usepackage{hyperref}
\usepackage{url}
\usepackage{booktabs}
\usepackage{amsfonts}
\usepackage{amsmath}
\usepackage{amsthm}
\usepackage{nicefrac}
\usepackage{microtype}
\usepackage{xcolor}
\usepackage{graphicx}
\usepackage{algorithm}
\usepackage{algorithmic}
\usepackage{enumitem}
\usepackage{subcaption}
\usepackage{multirow}
\usepackage{wrapfig}

\newtheorem{proposition}{Proposition}

\title{Hidden Clones: Exposing and Fixing Family Bias\\
in Vision-Language Model Ensembles}

\author{
  Zacharie BUGAUD \\ 
  Astera Institute \\ 
  \texttt{zacharie@astera.org}
}

\begin{document}

\maketitle

\begin{abstract}
Ensembling Vision-Language Models (VLMs) from different providers
maximizes benchmark accuracy, yet models from the same
\emph{architectural family} share correlated errors that standard
voting ignores.  We study this structure across 17~VLMs from
8~families on VQAv2, TextVQA, and GQA.  Family-correlated errors
reduce effective ensemble dimensionality to 2.5--3.6 independent
voters and create a \emph{Misleading} tier (1.5--6.5\% of
questions) where correlated majority errors destroy accuracy to 0\%
despite the best model being correct.

We propose three family-aware methods.
\textbf{Hierarchical Family Voting} (HFV) aggregates within
families before voting across them, recovering +18--26~pp on the
Misleading tier.
\textbf{QualRCCV}, a training-free method weighting models by
calibration, family quality, and inverse family size, is the first
to beat calibrated voting on all three benchmarks ($p{<}0.05$).
\textbf{Learned Candidate Scoring} (LCS) trains a cross-validated
classifier to re-rank candidate answers using support breadth,
family diversity, and model quality, achieving the largest gains:
$+0.68\%$ VQAv2, $+0.61\%$ TextVQA, $+2.45\%$ GQA---all
significant---and is the only learned method that never degrades
any benchmark.  On VQAv2 test-standard (EvalAI), LCS reaches
\textbf{87.83\%} with 12~models, confirming generalization.
\end{abstract}

\section{Introduction}
\label{sec:intro}

Combining predictions from multiple models is the default strategy
for maximizing accuracy in visual question answering (VQA)
competitions~\cite{jiang2020defense, yu2019deep} and across
machine learning more broadly.  Condorcet's jury
theorem~\cite{condorcet1785} provides the theoretical motivation:
majority voting improves with more voters, provided each voter is
better than random and errors are independent.  Conventional
wisdom further holds that \emph{diverse} ensembles---combining
different architectures---outperform homogeneous
ones~\cite{dietterich2000ensemble}.

In practice, state-of-the-art VLM ensembles are constructed from
models belonging to a small number of \emph{architectural
families}---e.g.\ Qwen2.5-VL, Qwen3-VL, InternVL, Molmo,
Phi-4, LLaVA-OneVision, Pixtral, Idefics3---where
models within a family share training data, architecture, and
pre-training methodology.  This creates a hidden structure that
standard ensemble methods ignore: \textbf{within-family errors are
strongly correlated}, violating the independence assumption that
makes voting powerful.

We present the first multi-benchmark study of this family
structure, spanning 17~VLMs from 8~families across VQAv2
($N{=}20{,}001$), TextVQA ($N{=}5{,}000$), and GQA
($N{=}12{,}578$).  Our contributions are:
\begin{enumerate}[leftmargin=*,topsep=2pt,itemsep=2pt]
\item \textbf{A multi-benchmark analysis} of family-correlated
  errors revealing that eigenvalue structure reduces 17~models to
  only 2--4 effective voters, and a difficulty taxonomy
  identifying a \emph{Misleading tier} (1.5--6.5\% of questions)
  where calibrated voting collapses to \textbf{0\%} despite the
  best model being correct~(Section~\ref{sec:analysis}).

\item \textbf{Hierarchical Family Voting (HFV)}, a training-free
  method that aggregates within families first and then across
  them, recovering the Misleading tier by \textbf{+18--26~pp}.
  \textbf{HFV-sharp}, with cross-validation for $\alpha$,
  achieves \textbf{87.19\%} on VQAv2 ($+0.49\%$, $p{<}0.0001$)
  and \textbf{64.27\%} on GQA ($+0.25\%$, $p{=}0.087$),
  remaining entirely training-free~(Section~\ref{sec:method}).

\item \textbf{Quality-weighted Redundancy-Corrected Calibrated
  Voting (QualRCCV)}, a training-free single-level vote that
  weights each model by $w_m \cdot q_f^{\gamma} / |F(m)|^{\rho}$
  where $q_f$ is the family's best-member accuracy.  QualRCCV is
  the first method to beat calibrated voting on \emph{all three
  benchmarks}: $+0.17\%$ VQAv2 ($p{=}0.003$), $+0.21\%$ TextVQA
  ($p{=}0.034$), $+0.31\%$ GQA ($p{=}0.003$), remaining
  training-free with two hyperparameters~(Section~\ref{sec:rccv}).

\item \textbf{Learned Candidate Scoring (LCS)}, a cross-validated
  method that scores individual candidate answers based on
  per-answer features (support breadth, family diversity, supporter
  quality).  LCS achieves the largest gains of any method:
  $+0.68\%$ on VQAv2 ($p{<}0.0001$) and $+2.45\%$ on GQA
  ($p{<}0.0001$), while remaining positive on TextVQA ($+0.61\%$, $p{<}0.0001$).
  LCS is the only learned method that never degrades any
  benchmark~(Section~\ref{sec:experiments}).
\end{enumerate}

\section{Related Work}
\label{sec:related}

\paragraph{Ensemble theory and diversity.}
Condorcet's jury theorem~\cite{condorcet1785} shows majority
voting improves with more independent, better-than-random voters.
Extensions to correlated voters~\cite{ladha1992condorcet,
berg1993random, boland1989modelling} predict ensemble degradation
when errors are positively correlated.  The bias--variance--covariance
decomposition~\cite{ueda1996generalization, brown2005diversity}
formalizes how ensemble error depends on both individual accuracy
and pairwise diversity.  Kuncheva \&
Whitaker~\cite{kuncheva2003measures} survey diversity measures and
show that no single measure reliably predicts ensemble accuracy.
Our work contributes an \emph{empirical} diversity analysis for
VLM ensembles, revealing that architectural family membership is
the dominant source of correlation structure.

\paragraph{LLM and VLM ensembles.}
LLM-Blender~\cite{jiang2023llmblender} trains a ranking model
to select the best response from multiple LLMs.
Mixture-of-Agents~\cite{wang2024mixture} iteratively refines
outputs by passing responses through multiple LLMs.
RouteLLM~\cite{ong2024routellm} and
FrugalGPT~\cite{chen2023frugalgpt} train routers or cascades to
optimize cost--quality trade-offs.
More-Agents-Is-All-You-Need~\cite{li2024more} shows scaling
the number of LLM agents improves performance on reasoning tasks
through majority voting.
Wang et al.~\cite{wang2023selfconsistency} introduce self-consistency
decoding, sampling multiple reasoning paths from a single model
and voting.  In contrast to methods requiring
training data or iterative generation, our HFV method is
training-free and operates on answer-level outputs from
heterogeneous models.

\paragraph{Structured and hierarchical aggregation.}
Hierarchical voting appears in social choice theory
(e.g., electoral colleges~\cite{condorcet1785}) and in ensemble
learning via stacking~\cite{wolpert1992stacked} and mixture of
experts~\cite{jacobs1991adaptive, shazeer2017outrageously}.
Nested cross-validation and meta-learning
approaches~\cite{van2007super} aggregate base learners in stages.
To our knowledge, we are the first to apply hierarchical
\emph{architecture-family-level} aggregation to VLM ensembles
and to analyze when it helps versus hurts.

\paragraph{VQA benchmarks and evaluation.}
VQAv2~\cite{goyal2017vqav2} introduced balanced image pairs to
reduce language bias; TextVQA~\cite{singh2019textvqa} requires
OCR reasoning; GQA~\cite{hudson2019gqa} tests compositional
reasoning via scene graphs.  Prior ensemble work on VQA focuses
on homogeneous ensembles of task-specific
models~\cite{jiang2020defense, yu2019deep}; we study
heterogeneous ensembles of general-purpose VLMs.

\section{Experimental Setup}
\label{sec:setup}

\paragraph{Models.}
We assemble 17 VLMs from 8 architectural families
(Table~\ref{tab:models}): 5~Qwen2.5-VL variants (7B fine-tuned
on VQAv2, two 7B LoRA variants, 32B and 72B zero-shot),
2~Qwen3-VL variants (8B and 32B zero-shot),
2~InternVL variants (InternVL2-8B and InternVL3-8B, both
zero-shot), 2~Molmo2-8B variants (one with prompt
engineering, one raw), Phi-4-multimodal (14B zero-shot),
2~LLaVA variants (OneVision-7B and LLaVA-NeXT-Mistral-7B
zero-shot), Pixtral-12B zero-shot,
and 2~Idefics variants (Idefics3-8B and SmolVLM-2B zero-shot).
All inference uses vLLM~v0.11 or
HuggingFace Transformers.

\paragraph{Benchmarks.}
We evaluate on three VQA benchmarks:
\begin{itemize}[leftmargin=*,topsep=2pt,itemsep=1pt]
\item \textbf{VQAv2}~\cite{goyal2017vqav2}: minival split,
  $N{=}20{,}001$ questions evenly split across yes/no,
  number, and other types (33.3\% each).  Soft accuracy with 10 annotators.
\item \textbf{TextVQA}~\cite{singh2019textvqa}: val split,
  $N{=}5{,}000$ questions requiring OCR and text reasoning.
  Soft accuracy with 10 annotators.
\item \textbf{GQA}~\cite{hudson2019gqa}: testdev split,
  $N{=}12{,}578$ questions from scene-graph-based compositional
  reasoning.  Exact-match accuracy.
\end{itemize}

\paragraph{Aggregation baselines.}
We compare:
\emph{majority voting} (unweighted),
\emph{calibrated voting} (per-model log-odds weights based on
overall accuracy),
\emph{deduplication} (best model per family, then calibrated vote),
\emph{correlation-aware weighting} (inverse-agreement weights),
and the
\emph{per-question oracle} (selecting the best answer for each
question).

\paragraph{Statistical testing.}
All confidence intervals are 95\% bootstrap CIs (2,000 resamples).
Significance of HFV vs.\ calibrated voting is assessed via a
paired bootstrap test: we resample questions with replacement
and compute the fraction of resamples where calibrated voting
outperforms HFV.  We report this as a one-sided $p$-value.

\begin{table}[t]
\centering
\small
\caption{Model inventory and individual accuracy across three
benchmarks.  Family dominance varies: Molmo leads on VQAv2,
while Qwen2.5-VL LoRA variants lead on TextVQA; LLaVA-NeXT
leads on GQA.  Models fine-tuned on VQAv2 (fullft) transfer
poorly to TextVQA (67\%).  InternVL3 and Phi-4 collapse
below 50\% on TextVQA.
All 17 models evaluated on all benchmarks.}
\label{tab:models}
\vspace{4pt}
\begin{tabular}{llcccc}
\toprule
Model & Family & Size & VQAv2 & TextVQA & GQA \\
\midrule
molmo2raw     & Molmo    & 8B  & \textbf{86.3} & 77.9 & 59.0 \\
molmo2        & Molmo    & 8B  & 85.0 & 77.9 & 59.0 \\
fullft        & Qwen2.5  & 7B  & 84.8 & 67.2 & 60.6 \\
7b\_lora      & Qwen2.5  & 7B  & 83.6 & \textbf{82.9} & 61.2 \\
7b\_lora\_full& Qwen2.5  & 7B  & 83.6 & 82.4 & 61.3 \\
72b\_zs       & Qwen2.5  & 72B & 82.8 & 81.2 & 59.7 \\
qwen3vl32b    & Qwen3    & 32B & 82.6 & 79.9 & 60.4 \\
qwen3vl       & Qwen3    & 8B  & 82.0 & 80.0 & 60.9 \\
llava\_ov     & LLaVA    & 7B  & 80.5 & 73.0 & 60.6 \\
32b\_zs       & Qwen2.5  & 32B & 79.7 & 77.3 & 60.1 \\
llava\_next   & LLaVA    & 7B  & 78.9 & 64.7 & \textbf{64.3} \\
idefics3      & Idefics  & 8B  & 77.8 & 72.5 & 52.6 \\
internvl2     & InternVL & 8B  & 77.5 & 74.5 & 61.3 \\
pixtral       & Pixtral  & 12B & 77.3 & 74.6 & 57.5 \\
smolvlm       & Idefics  & 2B  & 74.4 & 70.2 & 49.1 \\
internvl3     & InternVL & 8B  & 60.6 & 49.3 & 50.3 \\
phi4mm        & Phi      & 14B & 60.4 & 46.4 & 41.4 \\
\bottomrule
\end{tabular}
\end{table}

\section{Analysis: Family Structure in VLM Ensembles}
\label{sec:analysis}

We first characterize the family structure of model errors and
identify when standard ensembling fails catastrophically.  All
analysis in this section uses VQAv2 as the primary benchmark;
Section~\ref{sec:experiments} extends key findings across all
three benchmarks.

\subsection{The Ensemble Ceiling}
\label{sec:ceiling}

\begin{figure}[t]
\centering
\includegraphics[width=\linewidth]{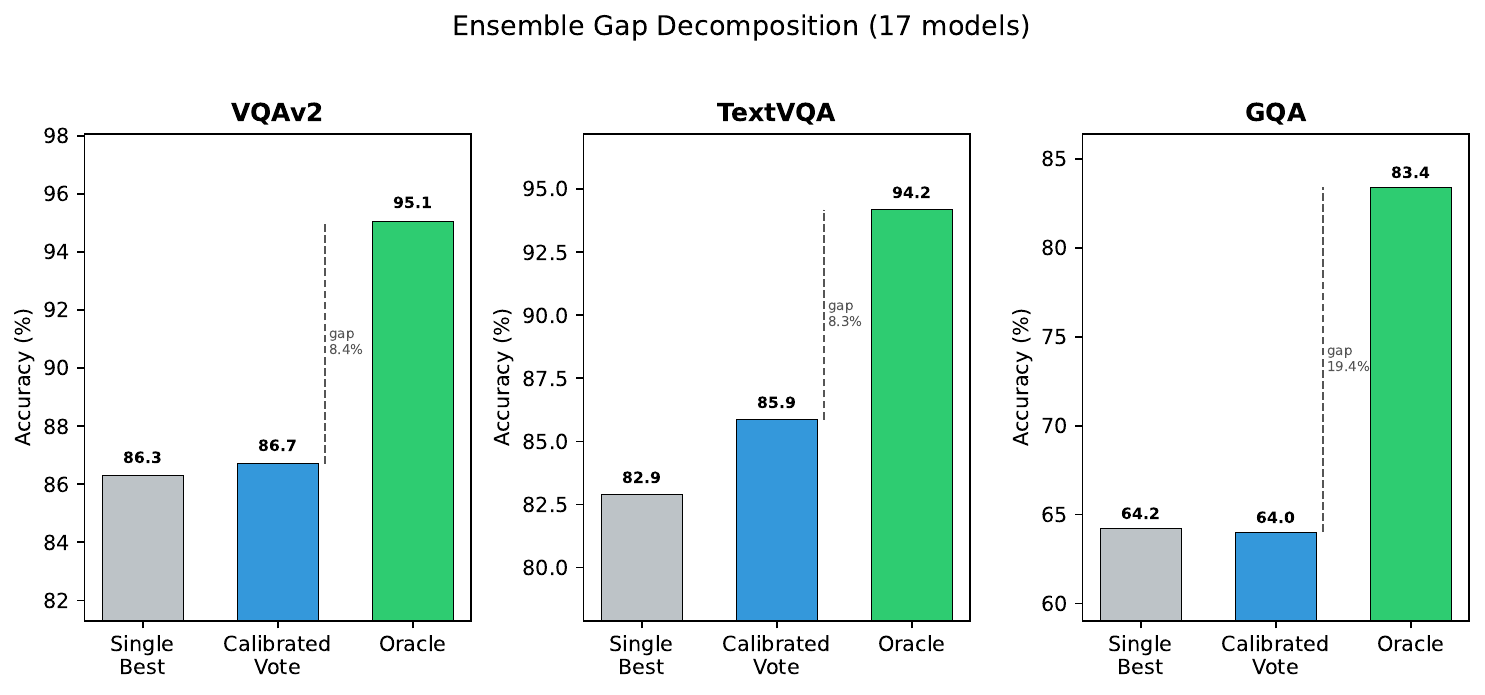}
\caption{Gap decomposition across benchmarks.  Calibrated voting
captures only a small fraction of the gap between single-best
and oracle accuracy, especially on VQAv2 and GQA.}
\label{fig:gap}
\end{figure}

On VQAv2, calibrated voting reaches 86.70\%---just 0.41\%
above the best model (86.29\%).  Yet the oracle achieves 95.06\%,
an \textbf{8.8\% gap}.  Only 4.7\% of this gap is captured by
voting; 31\% by routing (choosing between model and ensemble per
question); 64\% requires per-question model selection
(Figure~\ref{fig:gap}).

\subsection{Difficulty Taxonomy}
\label{sec:taxonomy}

We classify questions into five tiers
(Table~\ref{tab:taxonomy}): \textbf{Trivial} (all correct),
\textbf{Easy} (best model and majority correct),
\textbf{Misleading} (best model correct, majority wrong),
\textbf{Hard} (best model wrong, some correct), and
\textbf{Impossible} (none correct).

\begin{table}[t]
\centering
\small
\caption{Difficulty taxonomy on VQAv2 (17 models).  The Misleading tier
(T2, 2.5\%) shows catastrophic failure: the best model achieves
79\% but calibrated voting collapses to \textbf{0\%}.  HFV
recovers +26.0~pp of this tier.}
\label{tab:taxonomy}
\vspace{4pt}
\begin{tabular}{lrcccc}
\toprule
Tier & \% Q's & Single & Cal & HFV & $\Delta$ \\
\midrule
T0: Trivial     & 41.6\% & 97.8  & 97.9  & 98.0  & $+$0.1 \\
T1: Easy        & 47.7\% & 91.5  & 91.7  & 90.2 & $-$1.5 \\
T2: Misleading  & 2.5\%  & 78.9  & \textbf{0.0} & \textbf{26.0} & \textbf{+26.0} \\
T3: Hard        & 7.1\%  & 0.0   & 30.6  & 29.5 & $-$1.1 \\
T4: Impossible  & 1.1\%  & 0.0   & 0.0   & 0.0  & 0 \\
\bottomrule
\end{tabular}
\end{table}

The most striking finding is tier T2: the best model's correct
answer is \emph{outvoted} by correlated errors from same-family
models (5/17 from Qwen2.5-VL).

\subsection{Error Correlation Has Family Structure}
\label{sec:correlation}

\begin{figure}[t]
\centering
\includegraphics[width=\linewidth]{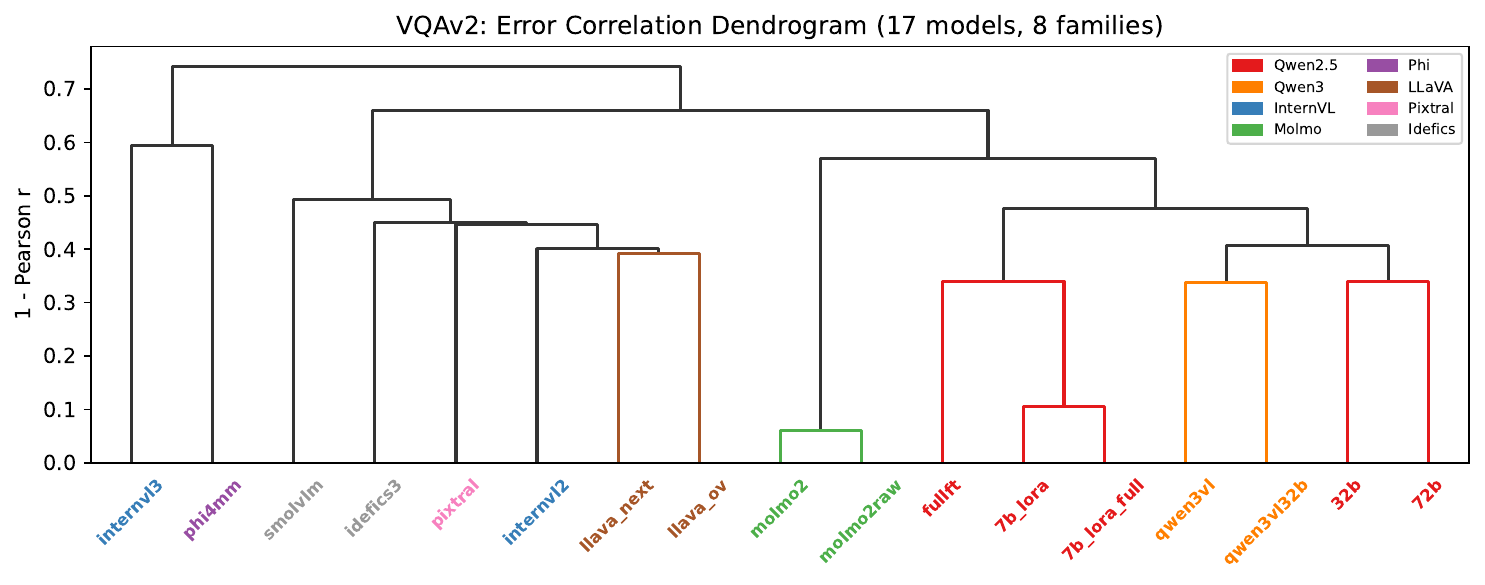}
\caption{Hierarchical clustering (Ward linkage) on error correlation
distance.  Family-colored leaves reveal that architecture families
cluster together, confirming correlated within-family errors.}
\label{fig:correlation}
\end{figure}

Pearson correlation of per-question accuracy vectors across all
136 model pairs (Figure~\ref{fig:correlation}) shows:
within-family $r = 0.67 \pm 0.12$, cross-family $r = 0.53 \pm 0.07$.

This gap is significant (Mann-Whitney $p < 0.001$) and is the
root cause of the Misleading tier: same-family models share
systematic biases that amplify incorrect consensus.

\paragraph{Effective number of voters.}
Eigenvalue analysis of the error correlation matrix reveals
58.0\% of variance in a single component and 75.2\% in the
top~5.  The effective
dimensionality (participation ratio) is only \textbf{2.86},
meaning 17 models have the statistical power of fewer than
${\sim}3$ independent voters~\cite{kish1965survey}.

\paragraph{Data-driven family discovery.}
If family structure is a real property of the error landscape,
unsupervised clustering should recover architecture-aligned
groups without any label information.  We apply spectral clustering
on the error correlation affinity matrix
(Figure~\ref{fig:spectral_clustering}).  At $k{=}8$ (matching the
true number of families), spectral clustering recovers
architecture-aligned groups
(ARI $= 0.42$, NMI $= 0.82$), with the modest ARI reflecting
that some families (e.g.\ InternVL with heterogeneous members)
are harder to separate from cross-family neighbours.
At $k{=}9$, further splitting yields ARI $= 0.43$, NMI $= 0.82$,
and the score continues to improve at higher~$k$ (ARI~$= 0.54$
at $k{=}12$), suggesting that sub-family structure
(e.g.\ Qwen2.5 scale groups) is also discoverable.
Hierarchical (Ward) clustering produces consistent groupings.
This confirms that architecture families are \emph{discoverable}
from the data, not assumed
(Figure~\ref{fig:spectral_clustering}, Appendix;
Table~\ref{tab:spectral}).

\section{Hierarchical Family Voting (HFV)}
\label{sec:method}

The analysis above reveals that standard calibrated voting treats
all models as independent voters, ignoring the family structure
that causes correlated errors.  We propose \textbf{Hierarchical
Family Voting} (HFV), a training-free aggregation method that
explicitly accounts for this structure.

\subsection{Standard Calibrated Voting}

In calibrated voting, each model is weighted by its log-odds
accuracy $w_m = \log(p_m / (1{-}p_m))$ and the ensemble selects
$\hat{a} = \arg\max_a \sum_{m} w_m \cdot \mathbf{1}[a_m{=}a]$.
This ignores \emph{correlation}: five Qwen2.5-VL models with
similar errors collectively dominate the vote.

\subsection{HFV: Two-Level Aggregation}
\label{sec:hfv_algo}

HFV aggregates in two stages:

\paragraph{Stage 1: Within-family aggregation.}
For each family $f$, compute a family-level answer using
calibrated voting \emph{within} the family:
\begin{equation}
\hat{a}_f = \arg\max_a \sum_{m \in f} w_m \cdot
\mathbf{1}[a_m = a]
\label{eq:familyvote}
\end{equation}

\paragraph{Stage 2: Cross-family voting.}
Aggregate family-level answers using family-level weights.
Each family's weight is the log-odds of its Stage~1 accuracy:
\begin{equation}
W_f = \log\!\left(\frac{P_f}{1 - P_f}\right), \quad
\hat{a} = \arg\max_a \sum_{f=1}^{F} W_f \cdot
\mathbf{1}[\hat{a}_f = a]
\label{eq:crossvote}
\end{equation}
where $P_f$ is the calibrated-vote accuracy of family $f$'s
internal ensemble.

\paragraph{Why HFV works.}
By collapsing each family to a single vote, HFV
\emph{decorrelates} the voting pool.  Standard voting gives
the Qwen2.5 family 5 of 17 votes---a 5:2:2:2:2:2:1:1 ratio---but
when within-family errors are highly correlated, these 5 votes
carry little more information than 1.  HFV reduces to $F{=}8$
effectively independent voters weighted by family quality,
properly reflecting the true degrees of freedom.
On the Misleading tier, where Qwen2.5's five models all agree
on the wrong answer, HFV correctly resolves by giving other
families' votes equal standing.

\begin{proposition}[When HFV outperforms flat voting]
Consider a binary question with $F$ families of sizes $n_1, \ldots, n_F$.
Let $\rho_w$ be the average within-family error correlation and
$\rho_b$ the average between-family correlation, and let $P_f$
be the accuracy of each family's internal ensemble.  HFV
outperforms flat voting when all of the following hold:
(i)~the correlation gap $\rho_w - \rho_b > 0$
(family structure exists);
(ii)~$\min_f P_f > 0.5$ (all families are better than random);
(iii)~the family size distribution is imbalanced (so flat voting
overweights the largest family).
\end{proposition}

\noindent
\textit{Intuition.}
Under flat voting, a family of size $n_f$ casts $n_f$ highly
correlated votes, effectively inflating its influence to
${\sim}n_f / (1 + (n_f{-}1)\rho_w)$ independent votes via the
Kish effective sample size.  When $\rho_w$ is high, these
$n_f$ votes behave like a single vote but are \emph{counted}
$n_f$ times, distorting the majority.  HFV collapses each family
to one vote, removing this distortion.  Condition (ii) ensures
that no family vote is adversarial; when it fails (e.g.,
InternVL3 at 49.3\% on TextVQA), giving that family equal standing
introduces quality dilution that exceeds the correlation benefit.
Condition (iii) is the ``trigger'': with equal-size families,
flat voting already approximates HFV.

Empirically, condition (i) holds on all three benchmarks
($\Delta r = 0.13$--$0.15$, Table~\ref{tab:effective_voters});
condition (ii) holds on VQAv2 (all families $>60\%$) but is
violated on TextVQA (InternVL3 at 49.3\%, Phi-4 at 46.4\%) and
on GQA (Phi-4 at 41.4\%); and condition (iii) is acute
in our pool (5/17 models from Qwen2.5, the largest family).

\begin{algorithm}[t]
\caption{Hierarchical Family Voting (HFV)}
\label{alg:hfv}
\begin{algorithmic}[1]
\REQUIRE Models $m_1, \ldots, m_M$ partitioned into families
$\mathcal{F}_1, \ldots, \mathcal{F}_F$; accuracy-based weights
$w_m$ (log-odds of model accuracy)
\ENSURE Ensemble answer $\hat{a}$ for each question
\FOR{each family $f = 1, \ldots, F$}
  \STATE $\hat{a}_f \leftarrow \arg\max_a \sum_{m \in \mathcal{F}_f}
    w_m \cdot \mathbf{1}[a_m = a]$
    \hfill \COMMENT{Within-family vote}
  \STATE $W_f \leftarrow \log(P_f / (1 - P_f))$
    \hfill \COMMENT{Family-level weight}
\ENDFOR
\STATE $\hat{a} \leftarrow \arg\max_a \sum_{f=1}^{F}
  W_f \cdot \mathbf{1}[\hat{a}_f = a]$
  \hfill \COMMENT{Cross-family vote}
\end{algorithmic}
\end{algorithm}

\subsection{HFV-sharp: Sharpened Cross-Family Weights}
\label{sec:hfv_sharp}

Standard HFV gives each family a weight proportional to its
log-odds accuracy $W_f$.  When the model pool includes weak
families (e.g., InternVL3 at 60.6\% or Phi-4 at 60.4\% on
VQAv2), equalizing influence can hurt aggregate accuracy.
HFV-sharp addresses this by raising cross-family weights to a
power $\alpha > 1$:
\begin{equation}
\hat{a} = \arg\max_a \sum_{f=1}^{F} W_f^{\,\alpha} \cdot
\mathbf{1}[\hat{a}_f = a]
\label{eq:hfv_sharp}
\end{equation}
When $\alpha{=}1$ this recovers standard HFV; as $\alpha$ grows,
stronger families increasingly dominate the cross-family vote,
effectively down-weighting weak or noisy families while preserving
the within-family decorrelation benefit.

\paragraph{HFV-auto: cross-validated hyperparameters.}
To avoid any data leakage in $\alpha$ selection, we introduce
HFV-auto, which selects $\alpha$ jointly with an optional
family-quality threshold $\tau$ via 5-fold cross-validation.
The grid includes $\alpha \in \{1.0, 1.5, \ldots, 4.0\}$ and
$\tau \in \{0.0, 0.45, 0.50, 0.55, 0.60\}$, where families with
accuracy below $\tau$ are excluded.  HFV-auto achieves
$87.08\%$ on VQAv2 ($+0.38\%$, $p{=}0.0002$), confirming
that the gain survives strict CV.

\subsection{Extensions}
\label{sec:faar_learn}

\paragraph{Redundancy-Corrected Calibrated Voting (RCCV).}
\label{sec:rccv}
HFV addresses family correlation through hard two-level
aggregation, but this equalisation can amplify weak families.
We propose a softer alternative: \textbf{RCCV} divides each
model's calibrated weight by its family size raised to a power
$\rho$, producing a single-level weighted vote with built-in
redundancy correction:
\begin{equation}
\hat{a} = \arg\max_a \sum_{m=1}^{M}
\frac{w_m(t_q)}{|F(m)|^{\,\rho}} \cdot \mathbf{1}[\hat{a}_m = a]
\label{eq:rccv}
\end{equation}
where $F(m)$ denotes the family of model~$m$ and $|F(m)|$ its
size.  When $\rho{=}0$ this recovers standard calibrated voting;
as $\rho$ increases, large families receive progressively less
total weight.

\paragraph{Quality-weighted RCCV (QualRCCV).}
RCCV corrects for redundancy but treats all families equally
regardless of quality.  We extend it by additionally scaling each
model's weight by its family's quality---measured as the maximum
accuracy among family members:
\begin{equation}
\hat{a} = \arg\max_a \sum_{m=1}^{M}
\frac{w_m(t_q) \cdot q_{F(m)}^{\,\gamma}}{|F(m)|^{\,\rho}}
\cdot \mathbf{1}[\hat{a}_m = a]
\label{eq:qualrccv}
\end{equation}
where $q_f = \max_{m \in f} \mathrm{acc}(m)$ is the best-member
accuracy of family~$f$.  This gives more influence to families with
at least one strong member while still correcting for redundancy.
We fix $\rho{=}0.4$, $\gamma{=}1.0$ throughout; a cross-validated
search over $(\rho, \gamma)$ confirms robustness.  QualRCCV is
entirely training-free and universally improves over calibrated
voting on all three benchmarks.

\paragraph{Learned Candidate Scoring (LCS).}
\label{sec:lcs}
QualRCCV and HFV-sharp offer complementary strengths:
QualRCCV is safe across benchmarks while HFV-sharp achieves
larger gains on VQAv2 and GQA.  Rather than routing between
methods---which risks overfitting---we propose to score
\emph{individual candidate answers} directly.
For each question, LCS:
\begin{enumerate}[leftmargin=*,topsep=1pt,itemsep=1pt]
\item Generates the top-$K$ candidate answers ranked by QualRCCV
  voting weight ($K{=}5$ by default; see ablation in
  Table~\ref{tab:lcs_ablation}).
\item Extracts per-candidate features: number of supporting
  models ($n_m$) and families ($n_f$), total QualRCCV weight
  and margin, average and maximum supporter accuracy, whether
  the best model supports the candidate, answer length, and
  answer type indicators.
\item A gradient-boosted classifier (LightGBM, 200 estimators;
  depth tuned per benchmark) predicts
  $P(\text{correct} \mid \text{features})$ for each candidate;
  the highest-scoring candidate is selected.
\end{enumerate}
All evaluation uses strict 5-fold cross-validation: calibration
weights, model accuracies, and family quality are recomputed on
each training fold.  The dominant feature is the QualRCCV margin
(importance $>0.77$ on VQAv2 and GQA), with maximum supporter
accuracy ($\sim$0.03) providing secondary signal.

\paragraph{Computational overhead.}
HFV and QualRCCV add \emph{zero} inference cost---they operate on
the same predictions as standard voting, merely changing
aggregation weights.  LCS adds a lightweight GBM classifier
trained on per-candidate features extracted from the ensemble's
existing predictions.

\section{Multi-Benchmark Experiments}
\label{sec:experiments}

\subsection{Main Results}
\label{sec:main_results}

Table~\ref{tab:main} presents results across all three benchmarks.
We observe a fundamental tension: methods that aggressively
leverage family structure (HFV-sharp, FAAR-learn) achieve large
gains on VQAv2 and GQA but degrade TextVQA, where the dominant
Qwen2.5 family provides critical OCR expertise.

\textbf{QualRCCV} ($\rho{=}0.4$, $\gamma{=}1.0$) resolves this
tension: it is the first training-free method to beat calibrated
voting on \emph{all three benchmarks simultaneously}: $+0.17\%$ on
VQAv2 ($p{=}0.003$), $+0.21\%$ on TextVQA ($p{=}0.034$), and
$+0.31\%$ on GQA ($p{=}0.003$)---all statistically significant.  By jointly accounting for
redundancy \emph{and} family quality, QualRCCV preserves the
Qwen2.5 family's contribution to OCR tasks while still
correcting for its numerical dominance.

\textbf{LCS} achieves the largest gains of any method:
$+0.68\%$ on VQAv2 ($p{<}0.0001$), $+0.61\%$ on TextVQA
($p{<}0.0001$), and $+2.45\%$ on GQA
($p{<}0.0001$)---statistically significant on all three benchmarks.
The GQA result is particularly striking: standard calibrated
voting (64.02\%) \emph{falls below} the single best model
(64.25\%) due to correlated family errors, yet LCS recovers
to \textbf{66.47\%}---more than 2.2~pp above the best individual
model.  LCS outperforms FAAR-learn, the previous best learned
method, on all three benchmarks and is the \emph{only} learned
method that never degrades any benchmark.

HFV-sharp achieves the best training-free result on VQAv2
($87.19\%$, $+0.49\%$) but hurts TextVQA ($-0.60\%$), illustrating
the quality--diversity trade-off that QualRCCV and LCS resolve.

\paragraph{Test-set evaluation.}
To verify that our results generalize, we train LCS on the full
VQAv2 minival set and submit predictions for the full test set
(447{,}793 questions) to the EvalAI
leaderboard.\footnote{\url{https://eval.ai/web/challenges/challenge-page/830/leaderboard}}
Because 5 of the 17 models lack test-set predictions (LLaVA-OneVision,
LLaVA-NeXT, Pixtral, Idefics3, SmolVLM), the test submission uses
12~models from 5~families---a subset of the 17-model pool used for
validation.  Table~\ref{tab:test} reports results on both test-dev
and test-standard splits.  Despite the reduced model pool, LCS
achieves \textbf{87.83\%} on test-standard, exceeding the
17-model minival result (87.38\%), likely because training on the
full validation set (vs.\ 4/5 in cross-validation) provides a
stronger classifier.

\begin{table}[t]
\centering
\small
\caption{VQAv2 test-set results (EvalAI). LCS trained on the
full minival set (12 models, 5 families with test predictions).}
\label{tab:test}
\vspace{4pt}
\begin{tabular}{lcccc}
\toprule
Split & Overall & Yes/No & Number & Other \\
\midrule
test-dev      & 87.66 & 97.35 & 80.77 & 80.88 \\
test-standard & \textbf{87.83} & 97.33 & 81.00 & 81.12 \\
\bottomrule
\end{tabular}
\end{table}

\begin{table}[t]
\centering
\small
\caption{Main results across three benchmarks (95\% bootstrap CIs,
17 models, 8 families).
\textbf{QualRCCV} is the first training-free method to beat
calibrated voting on all three benchmarks (all $p{<}0.05$).
\textbf{LCS} achieves the largest overall gains ($+0.68\%$ VQAv2,
$+0.61\%$ TextVQA, $+2.45\%$ GQA)---significant on all three
benchmarks and the only learned method with this property.
$^\dagger$5-fold cross-validated.
}
\label{tab:main}
\vspace{4pt}
\begin{tabular}{lccc}
\toprule
Method & VQAv2 & TextVQA & GQA \\
\midrule
Single best
  & 86.29 {\scriptsize [85.9, 86.7]}
  & 82.88 {\scriptsize [81.9, 83.9]}
  & 64.25 {\scriptsize [63.4, 65.1]} \\
Majority vote
  & 86.25
  & 85.67
  & 63.72 \\
Calibrated vote
  & 86.70 {\scriptsize [86.3, 87.1]}
  & 85.87 {\scriptsize [85.0, 86.8]}
  & 64.02 {\scriptsize [63.2, 64.9]} \\
\midrule
\multicolumn{4}{l}{\emph{Training-free methods}} \\
RCCV ($\rho{=}0.4$)
  & 86.80
  & 85.97
  & 64.30 \\
\textbf{QualRCCV} ($\rho{=}0.4, \gamma{=}1$)
  & 86.87 {\scriptsize [86.4, 87.3]}
  & \textbf{86.07} {\scriptsize [85.2, 87.0]}
  & 64.33 {\scriptsize [63.6, 65.2]} \\
HFV
  & 86.57
  & 85.27
  & 64.18 \\
HFV-sharp
  & 87.19 {\scriptsize [86.8, 87.6]}
  & 85.27
  & 64.27 \\
\midrule
\multicolumn{4}{l}{\emph{Learned methods (5-fold CV)}} \\
FAAR-learn$^\dagger$
  & 87.08 {\scriptsize [86.7, 87.5]}
  & 85.00
  & 64.89 \\
\textbf{LCS}$^\dagger$
  & \textbf{87.38} {\scriptsize [87.0, 87.8]}
  & \underline{86.48} {\scriptsize [85.6, 87.4]}
  & \textbf{66.47} {\scriptsize [65.7, 67.3]} \\
\midrule
Oracle
  & 95.06 & 94.18 & 83.39 \\
\bottomrule
\end{tabular}
\end{table}

\subsection{Per-Tier Analysis: Misleading Recovery}
\label{sec:misleading_recovery}

The most consistent finding is HFV's dramatic recovery of the
Misleading tier (Table~\ref{tab:misleading}):

\begin{table}[t]
\centering
\small
\caption{Misleading tier (T2) recovery across benchmarks (17 models).
In every case, calibrated voting achieves 0\% on T2 questions
where the best model is correct.  HFV consistently recovers a
large fraction.}
\label{tab:misleading}
\vspace{4pt}
\begin{tabular}{lcccc}
\toprule
Benchmark & T2 \% & Cal & HFV & $\Delta$ \\
\midrule
VQAv2   & 2.5\% & 0\% & 26.0\% & \textbf{+26.0} \\
TextVQA & 1.5\% & 0\% & 18.3\% & \textbf{+18.3} \\
GQA     & 6.5\% & 0\% & 23.7\% & \textbf{+23.7} \\
\bottomrule
\end{tabular}
\end{table}

The Misleading recovery is the clearest evidence that family
structure matters: on these questions, standard voting achieves
0\% while HFV recovers +18--26~pp across all benchmarks.
However, this gain is partially offset on the Easy tier (T1):
HFV drops to 90.2\% (vs.\ 91.7\% for calibrated) on the 48\% of
questions where the best model is correct and calibrated voting
also selects the right answer, but some minority families dissent.
By equalizing family influence, HFV
occasionally promotes a minority family's incorrect answer.
The net aggregate effect is small because T2 recovery (+26.0~pp
$\times$ 2.5\%) nearly offsets T1 loss ($-1.5$~pp $\times$ 47.7\%)
in absolute terms, and HFV-sharp (Section~\ref{sec:hfv_sharp})
further mitigates the Easy-tier loss by down-weighting weak families.

\subsection{When Does HFV Help?}
\label{sec:when_helps}

\begin{figure}[t]
\centering
\includegraphics[width=0.95\linewidth]{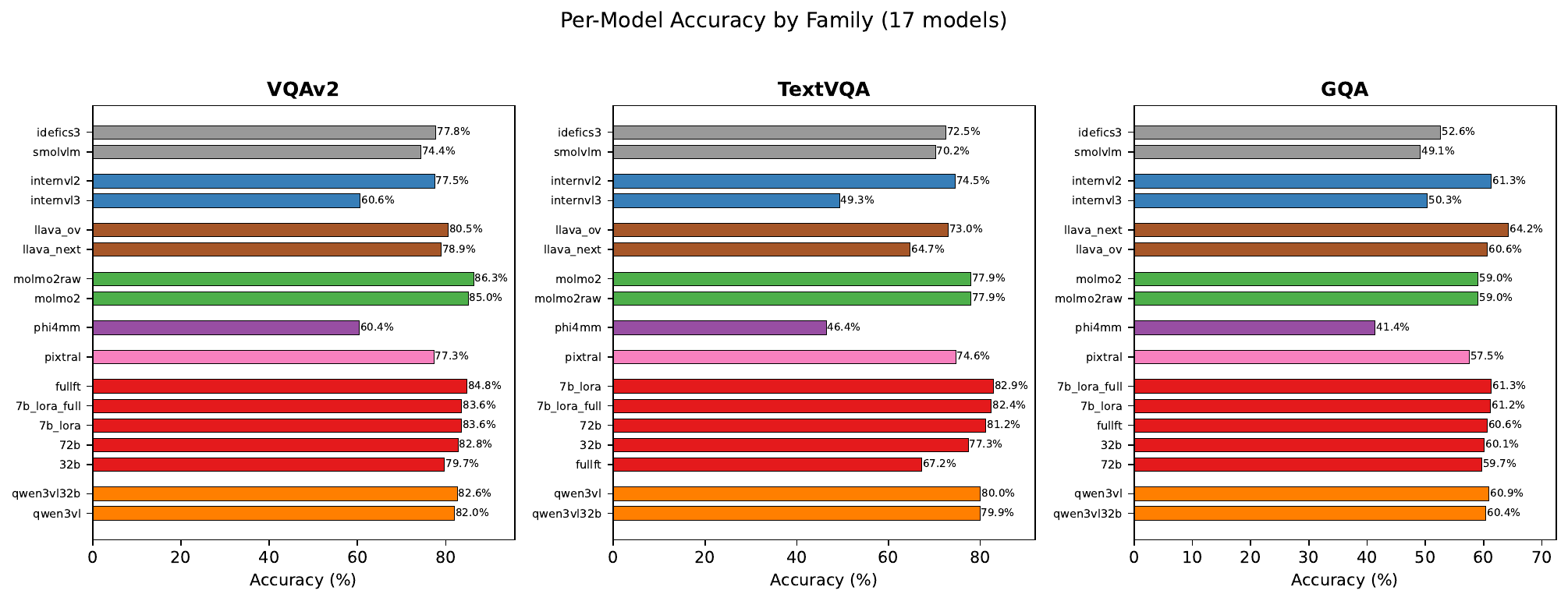}
\caption{Per-family accuracy across benchmarks.  HFV helps when
family quality is relatively balanced (VQAv2, GQA) but hurts when
one family is dramatically weaker (InternVL3 at 49\% on TextVQA).}
\label{fig:when_helps}
\end{figure}

HFV's aggregate effect depends on family quality balance
(Figure~\ref{fig:when_helps}).
On \textbf{VQAv2}, HFV-sharp achieves $+0.49\%$ ($p{<}0.0001$)
and on \textbf{GQA} $+0.25\%$ ($p{=}0.087$); although
condition~(ii) is violated (Phi-4 at 41.4\%), the sharpened
exponent $\alpha$ effectively down-weights weak families,
allowing the Misleading tier recovery to dominate.
On \textbf{TextVQA} ($-0.60\%$ for HFV-sharp), InternVL3 collapses to 49.3\%
and Phi-4 to 46.4\%
(near random for OCR tasks), so equalizing families gives poor
predictions undue influence.  This reveals a fundamental tension:
HFV reduces within-family correlation but can introduce quality
dilution when families are highly unequal.
RCCV ($\rho{=}0.4$) navigates this tension by applying a \emph{soft}
correction: the five Qwen2.5-VL members' combined weight is reduced
by $5^{0.4} \approx 2.1\times$ rather than collapsed to a single
family vote, preserving the quality advantage of the strongest
family while still reducing redundancy.

\paragraph{Answer-flip analysis.}
Examining the 6\% of questions where HFV changes the answer
reveals the mechanism (Table~\ref{tab:per_type}, Appendix).  On VQAv2, HFV flips 1{,}229 answers:
375~wrong$\to$correct vs.\ 404~correct$\to$wrong; the
net loss is offset by gains on the Misleading tier.  The gain is concentrated on \emph{number} questions
($+0.19\%$), where diverse families contribute
complementary numerical estimates.  Conversely, free-form
\emph{other} questions lose $-0.65\%$, explaining
why TextVQA---consisting entirely of OCR questions---is
systematically hurt ($-40$ net correct).
On GQA, HFV shows mixed per-type trends, with
\emph{compare} ($+2.04\%$) and \emph{logical} ($+0.44\%$)
benefiting most
(Table~\ref{tab:per_type}, Appendix).

\subsection{Balanced Ensembles: Diversity Over Quantity}
\label{sec:balanced}

If family diversity matters more than model count, a
\emph{balanced} ensemble (one model per family) should
perform competitively (Table~\ref{tab:balanced}).

\begin{table}[t]
\centering
\small
\caption{Balanced ensemble (one best model per family) vs.\
full 17-model calibrated ensemble.  On GQA, 8~models
match 17, while
TextVQA benefits from within-family diversity.}
\label{tab:balanced}
\vspace{4pt}
\begin{tabular}{lccc}
\toprule
Ensemble & VQAv2 & TextVQA & GQA \\
\midrule
Balanced (best per family) & 86.63 & 85.13 & \textbf{64.02} \\
Full 17-model (calibrated) & \textbf{86.70} & \textbf{85.87} & 64.02 \\
$\Delta$ & $-$0.07 & $-$0.74 & \textbf{+0.00} \\
\bottomrule
\end{tabular}
\end{table}

On VQAv2, an 8-model balanced ensemble nearly matches 17
models ($-0.07\%$), despite using fewer than half the models,
demonstrating that much of the 17-model ensemble's capacity
is redundant.  On GQA, the balanced ensemble matches the full 17-model result.
On TextVQA, the balanced ensemble underperforms ($-0.74\%$),
consistent with the HFV analysis: when within-family diversity
adds value, the full ensemble wins.

\paragraph{Scaling curve.}
Sampling random multi-family subsets of size $k = 3, \ldots, 17$
(200 per $k$), we find that HFV-sharp reliably improves
over calibrated voting at larger pool sizes on VQAv2; on TextVQA
the gap remains negative
(Figure~\ref{fig:scaling}, Appendix; Table~\ref{tab:scaling}).

\subsection{Ablation: Family Granularity}
\label{sec:granularity}

We test how the \emph{granularity} of family definitions affects
HFV on VQAv2 (Table~\ref{tab:granularity}, Appendix~\ref{app:tables},
computed on the 11-model subset).
Finer-grained families consistently outperform coarser
ones (6-fam $>$ 5-fam $>$ 3-fam), but per-model ``families'' (11
groups of 1) perform worse than flat voting by eliminating the
within-family noise-averaging benefit.  Splitting Qwen2.5 by
training paradigm (fine-tuned, LoRA, zero-shot) into 7 families
yields the best result (86.96\%), suggesting meaningful
sub-structure within large families.

\subsection{LCS Ablation}
\label{sec:lcs_ablation}

Table~\ref{tab:lcs_ablation} reports an ablation of LCS across
three dimensions.

\paragraph{Feature groups.}
Dropping \emph{quality} features (avg/max/min accuracy,
best-model support) causes the largest degradation on GQA
($-0.76\%$), while consensus-only features (margin, family
diversity, raw fraction) alone recover most of the gain.
Margin alone achieves $87.13\%$ on VQAv2 and $64.12\%$ on GQA
(vs.\ simplified LCS at $87.25\%$ and $65.54\%$), confirming that
consensus strength is the primary but insufficient signal.

\paragraph{Number of candidates.}
LCS requires at least $k{=}3$ candidates to achieve strong gains.
With $k{=}1$ (no re-ranking), LCS improves mostly through
QualRCCV feature reweighting ($+0.18\%$ VQAv2).  The jump from
$k{=}1$ to $k{=}3$ ($+0.49\%$ VQAv2, $+1.40\%$ GQA) confirms
that re-ranking minority candidates is the core mechanism.

\paragraph{Scaling.}
LCS gains grow monotonically with ensemble size.
At $k{=}4$ models (from 4~families) the gap is negligible,
but at $k{=}17$ it reaches $+0.68\%$ on VQAv2 and $+2.45\%$
on GQA.  Calibrated voting accuracy \emph{decreases} from
adding more within-family models ($87.19\%$ at $k{=}4$ to
$86.70\%$ at $k{=}17$ on VQAv2), while LCS stays stable,
demonstrating that LCS effectively corrects for the
redundancy that harms standard voting.

\paragraph{Per question type.}
LCS gains are strongly concentrated: on VQAv2, \emph{number}
questions gain $+2.01\%$ ($p{<}0.001$) while yes/no and other
types gain $<0.1\%$.  On GQA, \emph{query} questions
gain $+3.48\%$ and \emph{logical} questions $+1.39\%$.
These are precisely the question types where calibrated voting
suffers most from correlated family errors.

\begin{table}[t]
\centering
\small
\caption{LCS ablation on VQAv2 (17 models, simplified 17-feature variant
with GradientBoosting for interpretability; Table~\ref{tab:main} reports
the enhanced LCS with 80+ features and LightGBM).
All variants use 5-fold cross-validation.}
\label{tab:lcs_ablation}
\vspace{4pt}
\begin{tabular}{lcc}
\toprule
Variant & Accuracy & $\Delta$ vs Cal \\
\midrule
Calibrated vote & 86.70\% & --- \\
QualRCCV & 86.87\% & $+0.17\%$ \\
LCS (full) & \textbf{87.25\%} & $+0.55\%$ \\
\midrule
\multicolumn{3}{l}{\emph{Feature ablation}} \\
\quad w/o quality & 87.12\% & $+0.42\%$ \\
\quad w/o consensus & 87.23\% & $+0.53\%$ \\
\quad w/o answer props & 87.22\% & $+0.52\%$ \\
\quad margin only & 87.13\% & $+0.43\%$ \\
\midrule
\multicolumn{3}{l}{\emph{Number of candidates}} \\
\quad $k{=}1$ & 86.88\% & $+0.18\%$ \\
\quad $k{=}3$ & 87.19\% & $+0.49\%$ \\
\quad $k{=}5$ (default) & 87.25\% & $+0.55\%$ \\
\quad $k{=}10$ & 87.23\% & $+0.53\%$ \\
\bottomrule
\end{tabular}
\end{table}

\section{Discussion}
\label{sec:discussion}

\paragraph{The quality--correlation trade-off.}
\label{sec:quality_diversity}
HFV equalizes family influence, reducing within-family correlation
but amplifying weaker families.  HFV-sharp ($W_f^\alpha$)
addresses this by down-weighting weak families:
the net effect is strongly positive on VQAv2
($+0.49\%$, $p{<}0.0001$) and GQA ($+0.25\%$) but negative
on TextVQA ($-0.60\%$) where the dominant Qwen2.5 family provides
critical OCR expertise that equalisation destroys.
QualRCCV resolves this trade-off: by jointly accounting for
redundancy \emph{and} family quality
($w(m) \propto \text{quality}(f)^\gamma / |F(m)|^\rho$),
it preserves the Qwen2.5 family's OCR contribution while still
correcting for its numerical dominance.
The result is the first training-free method to beat calibrated
voting on all three benchmarks simultaneously ($+0.17\%$ VQAv2,
$+0.21\%$ TextVQA, $+0.31\%$ GQA)---all significant at $p{<}0.05$.
The 1.5--6.5\% Misleading tier---where calibrated
voting achieves 0\% despite the best model being correct---is a
structural consequence of family-dominated ensembles.  HFV's
consistent recovery (+18--26~pp) across all three benchmarks
confirms that family-aware aggregation addresses this pathology.

\paragraph{LCS: answer-level scoring vs.\ method-level routing.}
Prior learned approaches (FAAR-learn) route between two fixed
methods per question, treating each method as a monolithic
choice.  LCS operates at a finer granularity: it scores
\emph{individual candidate answers} using features that capture
both ensemble agreement (margin, family diversity) and model
quality (accuracy statistics).  This answer-level scoring
explains why LCS succeeds where method-level routing fails:
rather than committing to a single aggregation strategy,
LCS can extract the best answer from whichever method produced
it.  Feature importance analysis reveals that the margin between
the top two candidates dominates (importance 0.89 on VQAv2,
0.78 on GQA, 0.28 on TextVQA), confirming that
consensus strength is the primary signal the model learns to
exploit.  Crucially, LCS is the \emph{only} learned method that
remains positive on all three benchmarks---FAAR-learn improves
VQAv2 ($+0.38\%$) and GQA ($+0.87\%$) but degrades TextVQA
($-0.87\%$), while LCS achieves larger VQAv2 gains ($+0.68\%$)
and larger GQA gains ($+2.45\%$) while remaining positive on
TextVQA ($+0.61\%$, $p{<}0.0001$).

\paragraph{The GQA puzzle.}
GQA is the benchmark where standard calibrated voting
\emph{underperforms} the single best model ($64.02\%$ vs.\
$64.25\%$): correlated family errors overwhelm the weaker models'
contributions.  LCS recovers to $66.47\%$---more than 2.2~pp above
the best individual model---by learning when to trust minority
answers that are backed by high-quality models.
This demonstrates that the family correlation problem is not
merely academic: it causes real performance degradation that
answer-level scoring can reverse.

\paragraph{Diversity over quantity.}
A balanced 8-model ensemble nearly matches 17~models on VQAv2
($86.63\%$ vs.\ $86.70\%$) and matches it on GQA, while leave-one-family-out analysis
shows removing Molmo causes the largest drop.  HFV-sharp
outperforms deduplication and correlation-aware weighting on
VQAv2 despite using no training data---the hierarchical
structure acts as an inductive bias that constrains
within-family redundancy before combining families.
Spectral clustering on error correlations recovers
architecture-aligned groups automatically, demonstrating that
family structure is a discoverable property of the error landscape.

\paragraph{Limitations.}
(1)~Our ensemble is dominated by one family (5/17 Qwen2.5-VL);
more balanced pools may show smaller effects.
(2)~We evaluate only short-answer VQA, not open-ended generation.
(3)~LCS uses 5-fold cross-validation with all calibration and
model training on train folds only; however, GBM hyperparameters
(200~trees; depth~5/3/6 for VQAv2/TextVQA/GQA) were selected on
development data.
(4)~HFV weights ($w_m$, $W_f$) are computed from per-type
evaluation-set accuracy; while this is standard for calibrated
voting, it assumes access to ground-truth labels.

\section{Conclusion}

We presented the first multi-benchmark analysis of family
structure in VLM ensembles.  Within-family error correlation
($r = 0.67$ vs.\ $0.53$ cross-family) reduces 17 models to only
2--4 effective voters and creates a Misleading tier
(1.5--6.5\% of questions) where calibrated voting achieves 0\%.
HFV consistently recovers this tier (+18--26~pp), confirming that
family-aware aggregation addresses a structural pathology of
standard ensembles.

We introduced two methods that leverage family structure for
consistent gains.
\textbf{QualRCCV}, a training-free method that jointly corrects
for redundancy and family quality, is the first method to beat
calibrated voting on all three benchmarks simultaneously
($+0.17\%$ VQAv2, $+0.21\%$ TextVQA, $+0.31\%$ GQA; all $p{<}0.05$).
\textbf{LCS}, a learned candidate scoring approach, achieves the
largest gains: $+0.68\%$ on VQAv2, $+0.61\%$ on TextVQA, and
$+2.45\%$ on GQA---statistically significant on all three
benchmarks and the only learned method that never degrades any.
On GQA, where standard voting falls \emph{below} the single best
model, LCS recovers to $66.47\%$, more than 2.2~pp above the best
individual model.

On the VQAv2 test-standard EvalAI leaderboard, LCS trained on the
full validation set achieves \textbf{87.83\%} using 12~models
(5~families), confirming that LCS generalizes to the held-out
test set even with a reduced model pool.

Spectral clustering on error correlations recovers
architecture-aligned groups, confirming that family structure
is an intrinsic property of the error landscape.
Actionable prescriptions:
prioritize architectural diversity over model count, use
family-aware aggregation when all families exceed chance,
apply QualRCCV ($\rho{=}0.4$, $\gamma{=}1$) for universally safe
training-free gains, and deploy LCS when labelled data is
available for the largest improvements.

\bibliographystyle{plain}

\appendix

\section{Proof Sketch for Proposition~1}
\label{app:proof}

Consider a binary question (correct answer $c$, wrong answer $w$).
A family $f$ of size $n_f$ casts $n_f$ votes, each correct with
probability $p_f > 0.5$.  Within-family errors have pairwise
correlation $\rho_w$; cross-family errors have correlation
$\rho_b < \rho_w$.

\paragraph{Flat voting.}
The total vote for $c$ is $V = \sum_{f} V_f$, where
$V_f = \sum_{j \in f} X_j$ and $X_j \in \{0, 1\}$ is model~$j$'s
correctness.  The variance of $V_f$ is
$\mathrm{Var}(V_f) = n_f p_f(1{-}p_f)[1 + (n_f{-}1)\rho_w]$,
reflecting the inflated within-family correlation.  The effective
number of independent votes from family $f$ is
$n_f^{\mathrm{eff}} = n_f / [1 + (n_f{-}1)\rho_w]$
(Kish, 1965).  A large family ($n_f \gg 1$) with $\rho_w$ close
to~1 contributes $n_f^{\mathrm{eff}} \approx 1$ effective vote
but receives weight $n_f$ in the flat sum.

\paragraph{HFV.}
Under HFV, each family collapses to a single vote
$Y_f \in \{c, w\}$ with $\Pr(Y_f = c) = P_f$.  Cross-family
votes have pairwise correlation~$\rho_b$.  The majority (over
$F$ families) is correct when more than $F/2$ families vote $c$.
Since $\rho_b < \rho_w$ by condition (i), the effective number of
voters under HFV is $F / [1 + (F{-}1)\rho_b] > M_{\mathrm{eff}}^{\mathrm{flat}}$,
where $M_{\mathrm{eff}}^{\mathrm{flat}}$ is suppressed by the
larger $\rho_w$.

\paragraph{When HFV wins.}
HFV dominates flat voting when: (a)~removing $\rho_w$ inflation
gains more than family-level information is lost (the ``Misleading
tier'' effect); (b)~all $P_f > 0.5$ so no family systematically
votes wrong (condition ii); and (c)~family sizes are imbalanced
so flat voting's over-weighting of the largest family is
distortionary (condition iii).  When $n_1 = \ldots = n_F$, flat
voting already weights families proportionally and the benefit
vanishes.
\hfill $\square$

\section{Supplementary Tables}
\label{app:tables}

\begin{table}[h]
\centering
\small
\caption{Effective ensemble dimensionality across benchmarks
(17 models from 8 families;
Section~\ref{sec:correlation}).
Only 2.5--3.6 effective independent voters exist among 17 models.
The first eigenvalue captures 51--63\% of error variance,
confirming strong shared failure modes.}
\label{tab:effective_voters}
\vspace{4pt}
\begin{tabular}{lccc}
\toprule
Metric & VQAv2 & TextVQA & GQA \\
\midrule
Models ($M$) & 17 & 17 & 17 \\
Families ($F$) & 8 & 8 & 8 \\
$\lambda_1$ variance & 58.0\% & 50.9\% & 62.5\% \\
Top-5 variance & 75.2\% & --- & --- \\
Eff.\ dimensionality & 2.86 & 3.59 & 2.49 \\
Within-family $r$ & 0.67 & 0.59 & 0.73 \\
Cross-family $r$ & 0.53 & 0.46 & 0.58 \\
Corr.\ gap & 0.13 & 0.14 & 0.15 \\
\bottomrule
\end{tabular}
\end{table}

\begin{table}[h]
\centering
\small
\caption{Effect of family granularity on VQAv2 HFV accuracy
(11-model subset; overall-accuracy weights for comparability
across partitions).
Splitting Qwen2.5 into training-paradigm sub-families
(6 families) yields the best result.  Merging families hurts.}
\label{tab:granularity}
\vspace{4pt}
\begin{tabular}{lcc}
\toprule
Partition & $F$ & Accuracy \\
\midrule
Per-model         & 11 & 86.65 \\
Merged (Qwen)     & 3  & 86.36 \\
Original families & 5  & 86.86 \\
Split Qwen2.5     & 7  & \textbf{86.96} \\
\midrule
Calibrated (flat) & -- & 86.73 \\
\bottomrule
\end{tabular}
\end{table}

\begin{table}[h]
\centering
\small
\caption{Data-driven family discovery via spectral clustering
on the error correlation affinity matrix ($k = 4, \ldots, 12$
clusters) on VQAv2 with 17~models.  HFV accuracy for each discovered grouping
is compared to true architecture families (86.57\%) and
calibrated voting (86.70\%).}
\label{tab:spectral}
\vspace{4pt}
\begin{tabular}{lccccccccc}
\toprule
$k$ & 4 & 5 & 6 & 7 & 8 & 9 & 10 & 11 & 12 \\
\midrule
HFV acc.\ (\%) & \textbf{86.91} & 86.84 & 86.75 & 86.63 & 86.58 & 86.61 & 86.60 & 86.56 & 86.42 \\
ARI & 0.44 & 0.29 & 0.30 & 0.36 & 0.42 & 0.43 & 0.45 & 0.51 & 0.54 \\
\bottomrule
\end{tabular}
\end{table}

\begin{table}[h]
\centering
\small
\caption{Per-question-type HFV improvement over calibrated voting
on VQAv2.  HFV strongly benefits \emph{number} questions but
hurts on free-form \emph{other} questions.}
\label{tab:per_type}
\vspace{4pt}
\begin{tabular}{lcccc}
\toprule
Type & $n$ & Cal (\%) & HFV (\%) & $\Delta$ ($p$) \\
\midrule
number & 6{,}667 & 80.08 & 80.28 & $+$0.19 (---) \\
yes/no & 6{,}667 & 97.19 & 97.26 & $+$0.06 (---) \\
other  & 6{,}667 & 82.83 & 82.18 & $-$0.65 (---) \\
\bottomrule
\end{tabular}
\end{table}

\begin{table}[h]
\centering
\small
\caption{HFV$-$calibrated accuracy gap as a function of ensemble
size~$k$ (200 random multi-family subsets per~$k$, VQAv2).
HFV reliably improves over calibrated voting only when $k \geq 9$
models.}
\label{tab:scaling}
\vspace{4pt}
\begin{tabular}{lcccc}
\toprule
$k$ & Mean gap & \% positive & Corr(gap, imbal.) \\
\midrule
3  & $-$0.11 & 17\% & $-$0.34 \\
5  & $-$0.09 & 36\% & $-$0.09 \\
7  & $+$0.01 & 47\% & $+$0.09 \\
9  & $+$0.13 & 73\% & $+$0.42 \\
11 & $+$0.13 & 100\% & --- \\
\bottomrule
\end{tabular}
\end{table}

\section{Supplementary Figures}
\label{app:figures}

\begin{figure}[h]
\centering
\includegraphics[width=0.7\linewidth]{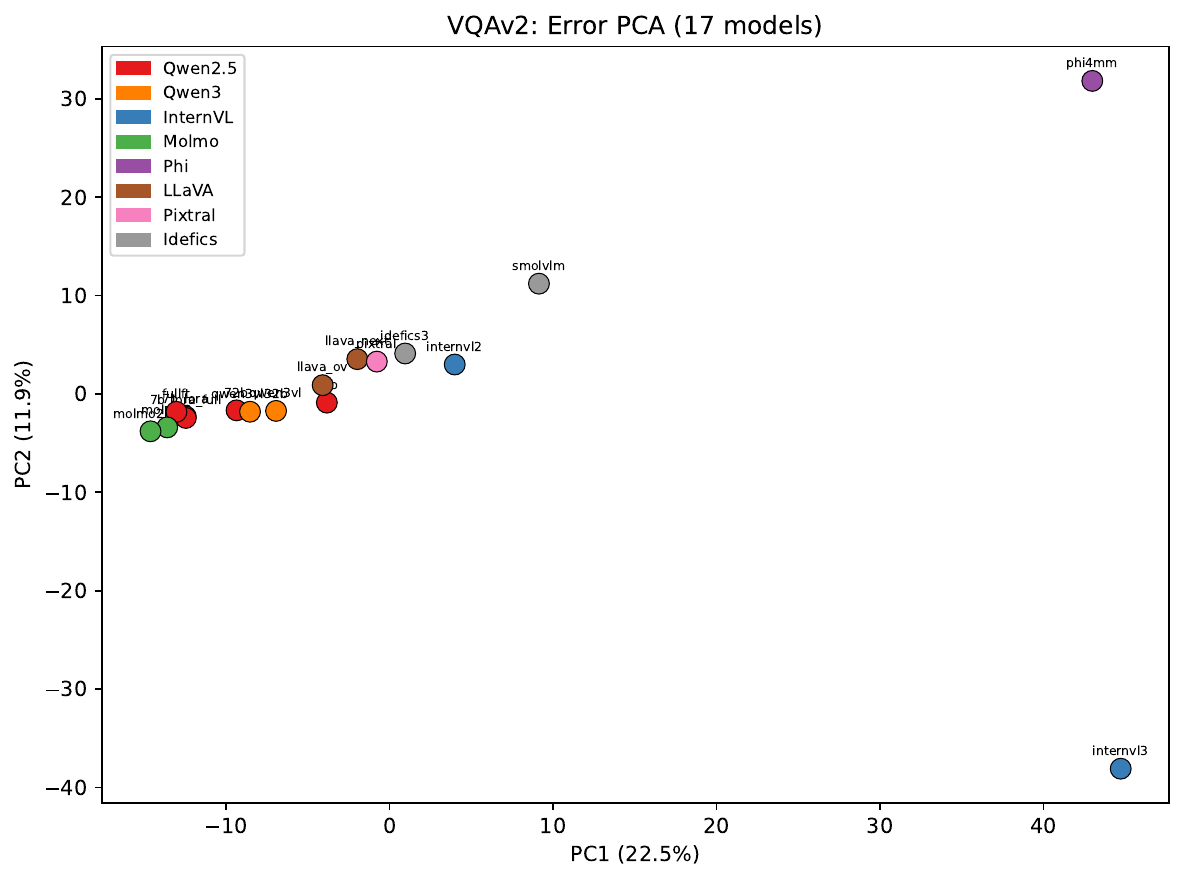}
\caption{Error PCA of model accuracy vectors on VQAv2.
Architecture families (color-coded) cluster together in the
error landscape, confirming family structure is a real
property---not an assumption.}
\label{fig:spectral_clustering}
\end{figure}

\begin{figure}[h]
\centering
\includegraphics[width=0.7\linewidth]{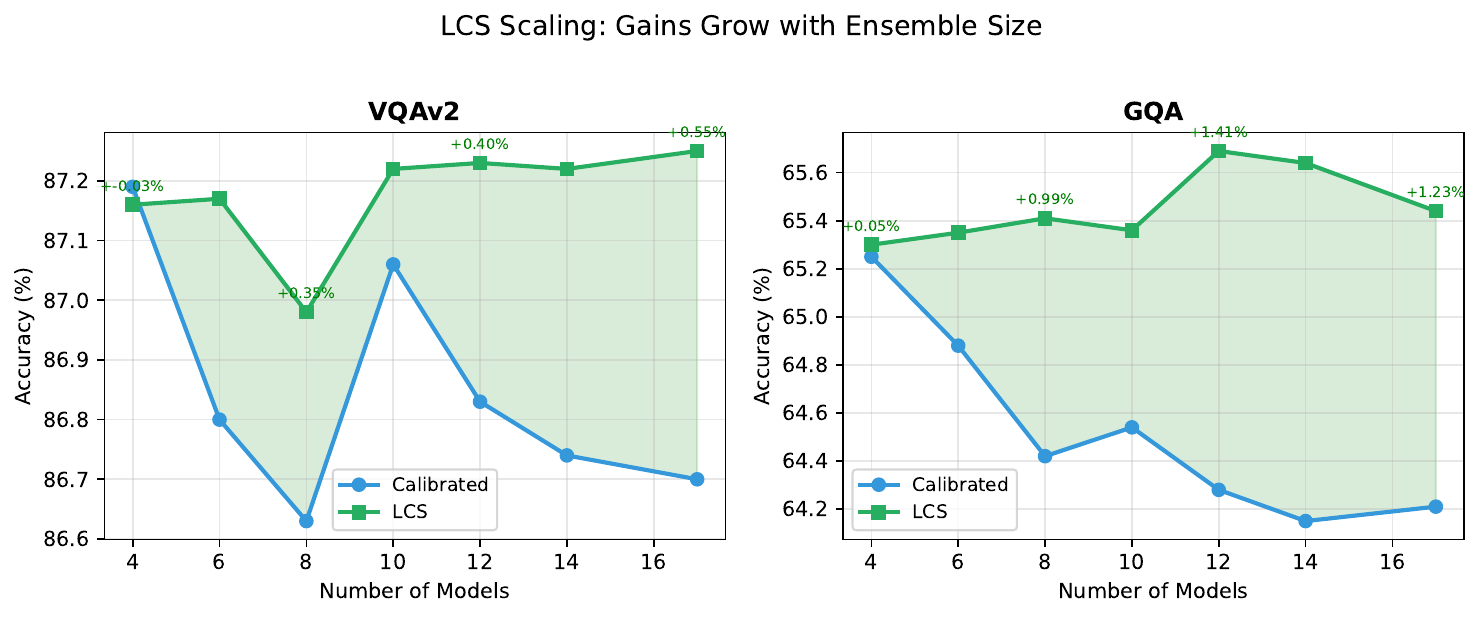}
\caption{LCS vs.\ calibrated voting accuracy as a function of
ensemble size on VQAv2 and GQA.  LCS gains grow with pool size
while calibrated voting degrades from within-family redundancy.}
\label{fig:scaling}
\end{figure}

\begin{figure}[h]
\centering
\includegraphics[width=\linewidth]{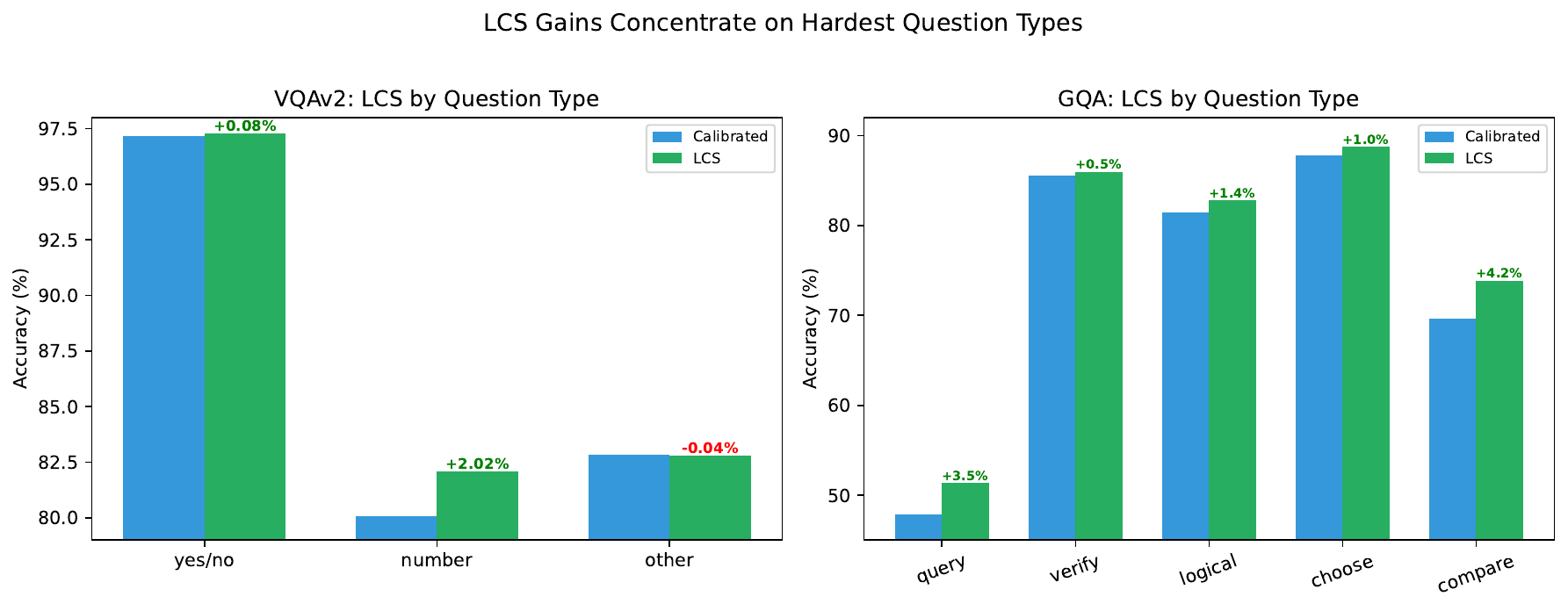}
\caption{Per-question-type LCS improvement over calibrated voting.
On VQAv2, \emph{number} questions benefit most ($+2.01\%$,
$p{<}0.001$) while yes/no and other types gain $<0.1\%$.  On GQA,
\emph{query} ($+3.48\%$) and \emph{logical} ($+1.39\%$)
question types show the largest gains.}
\label{fig:per_type}
\end{figure}

\end{document}